\documentclass[10pt,twocolumn,letterpaper]{article}

\usepackage{iccv}
\usepackage{times}
\usepackage{epsfig}
\usepackage{graphicx}
\usepackage{amsmath}
\usepackage{amssymb}

\usepackage{booktabs}
\usepackage{amsfonts}
\usepackage{multirow}
\usepackage[breaklinks=true,bookmarks=false]{hyperref}

\iccvfinalcopy 

\def\iccvPaperID{***} 

\ificcvfinal\fi

\begin{document}

\title{Virtual Accessory Try-On via Keypoint Hallucination}

\author{Junhong Gou\textsuperscript{1}, Bo Zhang\textsuperscript{1}, Li Niu\textsuperscript{1}, Jianfu Zhang\textsuperscript{1}, Jianlou Si\textsuperscript{2}, Chen Qian\textsuperscript{2}, Liqing Zhang\textsuperscript{1}\\
\textsuperscript{1}Shanghai Jiao Tong University, \textsuperscript{2}Sensetime Research\\
{\tt\small \{goujunhong, bo-zhang, ustcnewly, c.sis\}@sjtu.edu.cn, zhang-lq@cs.sjtu.edu.cn}\\
{\tt\small \{sijianlou, qianchen\}@sensetime.com}
}

\maketitle
\ificcvfinal\thispagestyle{empty}\fi

\begin{abstract}
   The virtual try-on task refers to fitting the clothes from one image onto another portrait image. In this paper, we focus on virtual accessory try-on, which fits accessory (\emph{e.g.}, glasses, ties) onto a face or portrait image. Unlike clothing try-on, which relies on human silhouette as guidance, accessory try-on warps the accessory into an appropriate location and shape to generate a plausible composite image. In contrast to previous try-on methods that treat foreground (\textit{i.e.}, accessories) and background (\textit{i.e.}, human faces or bodies) equally, we propose a background-oriented network to utilize the prior knowledge of human bodies and accessories. Specifically, our approach learns the human body priors and hallucinates the target locations of specified foreground keypoints in the background. Then our approach will inject foreground information with accessory priors into the background UNet. Based on the hallucinated target locations, the warping parameters are calculated to warp the foreground. Moreover, this background-oriented network can also easily incorporate auxiliary human face/body semantic segmentation supervision to further boost performance. Experiments conducted on STRAT dataset validate the effectiveness of our proposed method.
\end{abstract}

\section{Introduction}
Virtual try-on technology has gained popularity due to the booming online shopping industry. 
The goal of virtual clothing try-on technology is to improve the customer's online shopping experience by providing a sense of how the garment may look when worn.
Given an image with target clothes and another image with the target person, virtual clothing try-on could fit the target clothes onto the target person. Many previous works in this field have proposed flow-based~\cite{ge2021parser, han2019clothflow} or Thin Plain Spine (TPS) warping~\cite{han2018viton, minar2020cp, yang2020towards, wang2018toward, yang2021ct, ge2021disentangled} methods, which warped the clothes into shapes that fit the human body and then performed final synthesis. 

\begin{figure*}[t]
    \centering
    \includegraphics[width=0.99\linewidth]{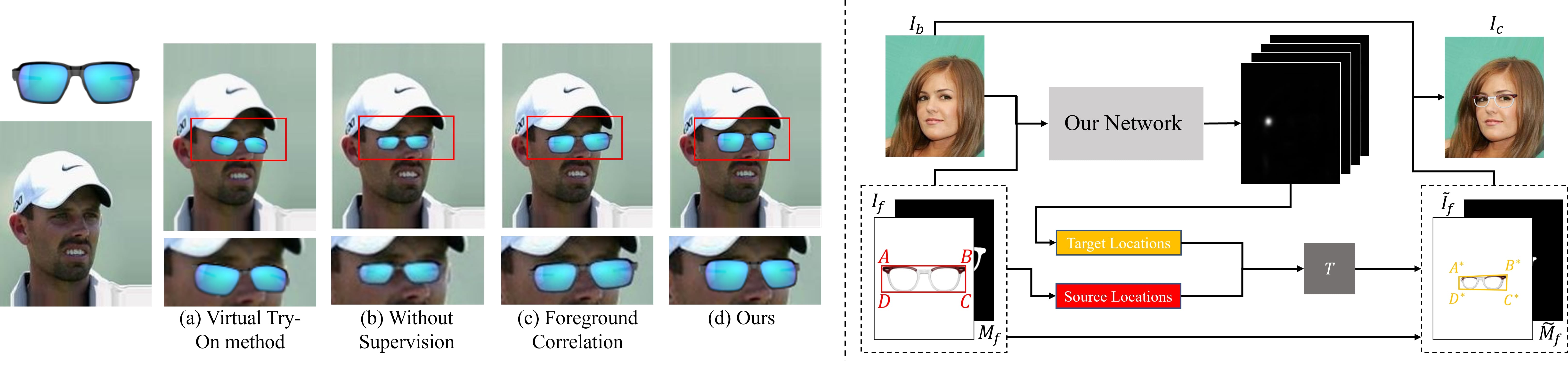}
    \caption{\textbf{On the left side}: synthesis results of a pair of foreground and background images with different methods. \textbf{On the right side}: the overall pipeline of our method, which can be described as follow. Given a background image $I_b$ and a foreground image $I_f$ with mask $M_f$, our network predicts four keypoint heatmaps, which indicate the target locations of four foreground keypoints ($A$, $B$, $C$, $D$) on the background image. Based on pairs of source and target locations, transformation matrix $T$ is calculated to warp the foreground image, which is combined with the background image to obtain the composite image $I_c$.}
    \label{fig:warp}
\end{figure*}

In this work, we focus on a specific and practical task: \textit{virtual accessory try-on} (\emph{e.g.}, glasses, hat, tie), which involves warping the foreground object (accessory) and pastes it on another background image (face or portrait image). 
The necessity of specifically designing methods for virtual accessory try-on is evident.
Simple cut-and-paste operations will inevitably bound to result in unrealistic synthesis results, like unnatural placement of foreground objects (\textit{e.g.}, position, scale) and mismatched camera viewpoints between foreground and background (\textit{i.e.}, human faces or human bodies). These issues could negatively affect the quality of the composite image. 
Although there are many previous virtual clothing try-on methods, they are unfortunately not appropriate for this accessory try-on task (See Figure \ref{fig:warp} (a)). 
The reason is virtual clothing try-on warps the target clothes to fit the human body, while virtual accessory try-on requires finding a suitable location and shape for the accessory without an explicit target to fit. Moreover, accessories can often be warped using rigid transformations, such as perspective transformations, leading to a more compact solution.

There are only a few works~\cite{azadi2020compositional, lin2018st, kikuchi2019regularized, li2021image, zhan2019spatial} focusing on virtual accessory try-on. They also fall into the scope of spatial transformation for image composition similar to existing virtual clothing try-on methods. 
In these works, the spatial transformation parameters are learned to warp the foreground, and the warped foreground and background are combined to create the composite images. 
Most of the previous works used generative models to synthesize composite images, but their performances are far from satisfactory (See Figure \ref{fig:warp} (b)). 
They fail to learn the matching of foreground and background due to the lack of supervision, \textit{i.e.}, the ground-truth composite image given a pair of foreground and background. 
Recently, Zhang \textit{et al.}~\cite{zhang2022spatial} constructed a dataset named STRAT dataset with ground-truth supervision, which contains three sub-datasets (STRAT-glasses, STRAT-hat, and STRAT-tie).
To take advantage of ground-truth supervision, CorrelNet is proposed to capture the correspondence between foreground pixels and background pixels, based on which the warping parameters could be calculated to warp the foreground. 
However, capturing the correlations between foreground and background pixels is not an easy task due to the significant appearance difference between the foreground and the background. The results of CorrelNet are still not promising (See Figure \ref{fig:warp} (c)).
Intuitively, the foreground may contain noisy information (\textit{e.g.}, different colors, texture), which is harmful to predicting the target locations, while the background information can determine the approximate target locations of foreground keypoints based on the \textit{prior knowledge of human body} (\textit{e.g.}, a tie should be around the neck, or the legs of the glasses should be on the ears). The foreground information with \textit{prior knowledge of accessory} (\textit{e.g.}, shape, fine-grained category) can refine location prediction and transformation matrix based on results estimated by background.
Like the situation in Fig.\ref{fig:warp}, the man is slightly facing to the right, so the glasses should also be facing to the right based on human body priors. Then based on accessory priors, the glass on the left should look larger than the right one. 
Hence, we argue that the background should play a more important role than the foreground.

Motivated by the above point, we design a background-oriented network to learn human body priors and hallucinate background keypoints, \emph{i.e.}, the target locations (four vertices of the foreground bounding box) on the background.
Our hallucination network mainly considers the background image. We adopt UNet~\cite{ronneberger2015u} structure, which takes in the background image to predict heatmaps for each of the four target locations.
We employ an encoder to extract foreground information with accessory priors, which is then injected into the bottleneck of UNet using a Dual Attention Fusion (DAF) module.
According to the predicted target locations, we can calculate a transformation matrix $T$ to warp the foreground, combined with the background to achieve the composite image.
Our background-oriented network design will leverage extra information (\textit{e.g.}, semantic segmentation) of background, which can be easily incorporated by employing an additional decoder. 
The decoder features with rich semantic information are added to the original decoder features, helping predict the target locations. Note that the semantic information is only used as auxiliary information during training.
Our proposed method is effective and flexible. It can fast adapt to new accessories in these categories without additional training.
To evaluate our proposed method, we conduct extensive experiments on the STRAT dataset \cite{zhang2022spatial} and compare it with previous works, which proves that our method can achieve excellent performance. 

\section{Related Work}

\paragraph{Virtual Try-on}
The popularity of online shopping has made virtual try-on essential for consumers to improve their online shopping experience. Referring to~\cite{feng2022weakly, he2022style}, we can divide the existing virtual try-on technologies into 2D and 3D categories. 3D virtual try-on technology can bring a better user experience, but it relies on 3D parametric human models and building a large-scale 3D dataset for training is expensive. Compared with 3D, image-based virtual try-on, that is, 2D virtual try-on, although not as flexible as 3D (\textit{e.g.}, allowing being viewed with arbitrary views and poses), is more light-weighted and generally more prevalent.

Many previous 2D virtual try-on work~\cite{han2018viton, wang2018toward, yang2020towards, zheng2019virtually, minar2020cp, ge2021disentangled} have used the Thin Plain Spine (TPS) method to flexibly deform clothes to cover the human body. However, TPS can only provide simple deformation processing, which can only roughly migrate the clothing to the target area and cannot handle some larger geometric deformations. More recently, CT-Net~\cite{yang2021ct} combined distance field guided dense warping and TPS warping to achieve more precise clothing transfer. In addition, many flow-based methods~\cite{han2019clothflow, ge2021parser, he2022style, bai2022single} have been proposed, they modeled the appearance flow field between clothes and corresponding regions of the human body to better fit the clothes to the person. There are also some methods~\cite{choi2021viton, lee2022high} to deal with the virtual try-on task under the high-resolution image, which undoubtedly has higher quality requirements in the warping of clothes and the synthesis of images.

The 2D virtual try-on works mentioned above are mainly for the try-on of clothes, while we focus on the try-on of accessories such as glasses, hat, and tie. There are only a few works \cite{azadi2020compositional, lin2018st, kikuchi2019regularized, li2021image, zhang2022spatial, zhan2019spatial} on virtual accessory try-on, which are also covered by spatial transformation for image composition and will be discussed in the next subsection. 

\paragraph{Image Composition}
Image composition means overlaying a foreground object on a background image to generate a composite image. However, the obtained composite image may look unrealistic and the unreality could be caused by many issues.
One issue is the different lighting conditions between foreground and background. Image harmonization~\cite{cong2021bargainnet, cong2020dovenet, tsai2017deep, ling2021region, zhu2015learning} sought to fix this issue by correcting the illumination statistics of the foreground to make it compatible with the background, resulting in a harmonious composite image. 
Another issue is the unreasonable placement of the foreground object. To find reasonable locations and scales to place the foreground object, many object placement methods~\cite{dvornik2018modeling, lee2018context, zhang2020and, zhang2020learning} have been proposed. To further support more flexible shape adjustment, spatial transformation methods~\cite{azadi2020compositional, lin2018st, kikuchi2019regularized, li2021image, zhang2022spatial, chen2019toward, zhan2019spatial} for image composition predicted the warping parameters of the foreground object. By taking virtual accessory try-on as an example application, previous works~\cite{azadi2020compositional, lin2018st, kikuchi2019regularized, li2021image, zhan2019spatial} studied spatial transformation for the accessory image to obtain the composite image. Most of them concatenated foreground and background as input to predict the warping parameters. Due to the lack
of ground-truth supervision (the ground-truth composite image given a pair of foreground and background), most of the previous works used adversarial learning to generate composite images which are indistinguishable from real images, but their performances are far from satisfactory. The recent work~\cite{zhang2022spatial} modeled the correlation between foreground and background, calculated the warping parameters on this basis, and achieved good results using the ground-truth supervision in their proposed STRAT dataset. All the above methods treat foreground and background equally. In contrast, we propose a novel background-oriented network which is more suitable for virtual accessory try-on.

\begin{figure*}[!t]
    \centering
    \includegraphics[width=1.0\linewidth]{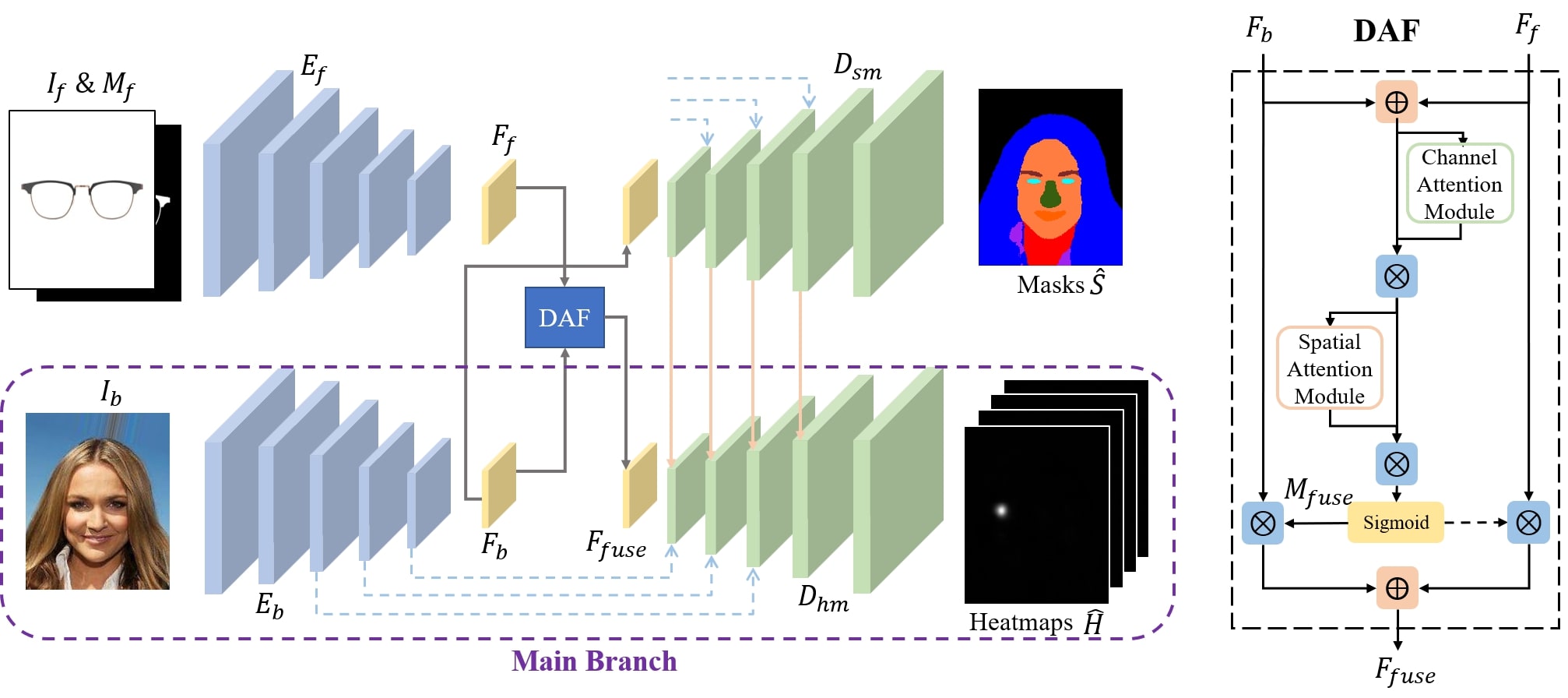}
    \caption{Overview of our proposed network structure. The main branch is a UNet with background encoder $E_b$ and heatmap decoder $D_{hm}$, which takes in background $I_b$ and predicts four heatmaps $\hat{H}$. The foreground with mask $\{I_f, M_f\}$  passes through $E_f$ and the output foreground feature map $F_f$ is fused with $F_b$ by our DAF module (detailed structure shown in the right subfigure). The semantic decoder $D_{sm}$ predicts background semantic masks $\hat{S}$ and the decoder features are added to the heatmap decoder $D_{hm}$ to help predict heatmaps. The blue dashed lines denote the skip links between $E_b$ and $\{D_{hm}, D_{sm}\}$, while the orange solid lines denote the connection between two decoders.}
    \label{fig:model}
\end{figure*}

\section{Our Method}

The overall view of our model is shown in Figure \ref{fig:model}. The main branch is background UNet~\cite{ronneberger2015u}, which takes in background image $I_b$ and produces four heatmaps corresponding to the target locations of four foreground keypoints. The background UNet consists of a background encoder $E_b$ and a heatmap decoder $D_{hm}$. We adopt ResNet18~\cite{he2016deep} as $E_b$ to extracts the background feature map $F_b$.  The decoder $D_{hm}$ adopts the decoder structure in~\cite{li2020natural}. 
We use the foreground encoder $E_f$ with the same structure as $E_b$ to extract the foreground feature map $F_f$.  We inject foreground information into the main branch, by fusing $F_f$ and $F_b$ through our Dual Attention Fusion (DAF) module. To further improve the performance of our network, we introduce a semantic decoder $D_{sm}$ which takes in the background feature map $F_b$ and predicts background semantic masks. The semantic decoder $D_{sm}$ shares the same structure as the heatmap decoder $D_{hm}$. 
Next, we will first introduce our main branch for heatmap prediction. 

\subsection{Main Branch}
As the main branch of our network, background UNet contains a background encoder $E_b$ and a heatmap decoder $D_{hm}$, and is able to generate heatmaps $\hat{H}$ based on the input background image $I_b$. Following the structure in~\cite{li2020natural}, the decoder $D_{hm}$ consists of five stages, where the first four stages consist of two to three stacked residual blocks with a deconvolution layer included in the first residual block of each stage, and the last stage consists of only a deconvolution layer and a convolutional layer to obtain the final output. The encoder feature maps from the last three stages in the encoder $E_b$ are added to the feature maps in the corresponding stages in the decoder $D_{hm}$ via skip connection. We denote the predicted (\emph{resp.}, ground-truth) $k$-th heatmap as $\hat{H}_k$ (\emph{resp.}, $H_k$). We omit the subscript $k$ when not specifying a heatmap. 

\paragraph{Loss Function}
Instead of simply setting the value at the target location to $1$ and the values at other locations as $0$, the ground-truth heatmap is generated by plotting a Gaussian distribution with radius $g$ centered at the target location. Directly labeling the locations close to the target location as $0$ could hinder the training of the network because they are highly similar to the target location. With the soft ground-truth heatmap, we observe that the network is easier to converge. 

We employ Adaptive Wing (AWing) Loss~\cite{wang2019adaptive} to measure the difference between predicted heatmaps and ground-truth heatmaps, which is defined as follows,
\begin{equation}
    AWing(y, \hat{y})= 
    \begin{cases}
    \omega\ln{(1+|\frac{y-\hat{y}}{\epsilon}|^{\alpha-y})},&{\text{if }}\ |y-\hat{y}| < \theta,\\
    A|y-\hat{y}|-B,&{\text{otherwise,}}
    \end{cases}
\end{equation}
where $y$ and $\hat{y}$ denote the value at a certain location on the ground-truth and predicted heatmap respectively. $\omega$, $\alpha$, $\theta$, and $\epsilon$ are positive hyper-parameters. $A=\omega(1/(1+(\theta/\epsilon)^{(\alpha-y)}))(\alpha-y)((\theta/\epsilon)^{(\alpha-y-1)})(1/\epsilon)$ and $B=(\theta A - \omega\ln{(1+(\theta/\epsilon)^{(\alpha-y)}}))$ make the loss function continuous and smooth at $|y-\hat{y}|=\theta$. Following~\cite{wang2019adaptive}, we set $\alpha=2.1$, $\omega=14$, $\epsilon=1$, $\theta=0.5$. Compared with commonly used MSE loss, AWing loss is more sensitive to some subtle errors, with a higher gradient when $|y-\hat{y}|$ approaches 0. 

In addition, the number of positive locations (locations with large values) and negative locations (locations with small values) in the ground-truth heatmap are highly unbalanced. To address this issue, we tend to assign larger weights to the positive locations when calculating the loss. Following~\cite{wang2019adaptive}, we define a positive mask $M_{pos}$ to indicate the positive locations:
\begin{equation}
    M_{pos}(p) = \begin{cases}
        1,& {\text{where }} \tilde{H}(p) \geq 0.2, \\
        0,& {\text{otherwise,}}
    \end{cases}
\end{equation}
where $p$ is a location, $\tilde{H}$ is generated from ground-truth heatmap $H$ by $3 \times 3$ dilation, $M_{pos}(p)$ (\emph{resp.}, $\tilde{H}(p)$) is the value at location $p$ on $M_{pos}$ (\emph{resp.}, $\tilde{H}$).
Intuitively, the positive locations include the target location and a part of the surrounding locations.

With the positive mask $M_{pos}$, the heatmap prediction loss $\mathcal{L}_{hm}$ is calculated as follows,
\begin{equation}\label{eq:wm}
    \mathcal{L}_{hm}=AWing(H, \hat{H})\otimes (\gamma \cdot M_{pos} + 1),
\end{equation}
where $\gamma$ is a hyper-parameter to control the degree of magnification and $\otimes$ means element-wise multiplication.

\paragraph{Spatial Transformation}
After obtaining the heatmaps, we do not directly take the location with the maximum value in each heatmap as the predicted target location. For robustness, we use the values in the heatmap as weights to calculate the target location. Similar to~\cite{sun2018integral}, given the predicted $k$-th heatmap $\hat{H}_k$, the $k$-th target location $v^t_k$ is obtained by
\begin{equation}
    v^t_k = \sum_{p=1}^N p\cdot softmax(\beta 
    \cdot \hat{H}_k(p)),
\end{equation}
in which $N$ is the size of heatmap, $p$ is a location, and $\hat{H}_k(p)$ is the value at location $p$ on $\hat{H}_k$. 
$\beta$ is a hyper-parameter (set as 1000) to exaggerate the impact of locations with near-maximum values and $softmax$ means softmax normalization. 

The perspective transformation matrix $T$ is calculated according to the target locations $\{v^t_k|_{k=1}^4\}$ and their source locations  $\{v^s_k|_{k=1}^4\}$ on the foreground image. The details of calculating the transformation matrix based on pairs of source and target locations can be found in~\cite{zhang2022spatial}. Then, we warp the foreground image according to the perspective transformation matrix $T$ to get the warped foreground $\tilde{I}_f$ and the warped foreground mask $\tilde{M}_f$. Finally, the composite image $I_c$ can be obtained by
\begin{equation}
    I_c = \tilde{M}_f\otimes \tilde{I}_f + (1-\tilde{M}_f)\otimes I_b.
\end{equation}
This process is shown in Figure \ref{fig:warp}.

\subsection{Injecting Foreground Information}

\begin{table*}[!ht]
    \centering
    \resizebox{\textwidth}{!}{
    \begin{tabular}{l|cccc|cccc|cccc}
        \hline
        \multirow{2}*{Method} & 
         \multicolumn{4}{c}{ STRAT-glasses } & \multicolumn{4}{|c|}{ STRAT-hat } & \multicolumn{4}{c}{ STRAT-tie }  \\
         & LSSIM$\uparrow$ & IoU$\uparrow$ & Disp$\downarrow$ & User$\uparrow$ & LSSIM$\uparrow$ & IoU$\uparrow$ & Disp$\downarrow$ & User$\uparrow$ & LSSIM$\uparrow$ & IoU$\uparrow$ & Disp$\downarrow$ & User$\uparrow$\\
         \hline\hline
         ST-GAN & 0.5655 & 0.5932 & 0.0240 & 4.33\% & 0.4362 & 0.6859 & 0.0316 & 1.80\% & 0.2780 & 0.1126 & 0.0440 & 0.80\% \\
         ST-GAN(+s) & 0.6061 & 0.6579 & 0.0198 & 0.87\% & 0.5164 & 0.7455 & 0.0235 & 4.80\% & 0.2517 & 0.1211 & 0.0393 & 0.53\% \\
         CompGAN & 0.5362 & 0.5593 & 0.0279 & 4.27\% & 0.4311 & 0.6675 & 0.0343 & 3.80\% & 0.2768 & 0.0918 & 0.0506 & 0.20\% \\
         CompGAN(+s) & 0.5807 & 0.6353 & 0.0216 & 2.00\% & 0.5047 & 0.7303 & 0.0246 & 1.87\% & 0.2452 & 0.1064 & 0.0425 & 0.40\% \\
         RegGAN & 0.5356 & 0.5069 & 0.0299 & 1.67\% & 0.4028 & 0.6147 & 0.0371 & 1.13\% & 0.2469 & 0.0792 & 0.0603 & 0.93\% \\
         SF-GAN & 0.5406 & 0.5472 & 0.0267 & 2.27\% & 0.4140 & 0.6521 & 0.0365 & 1.87\% & 0.2575 & 0.0885 & 0.0544 & 0.33\% \\
         AGCP & 0.5240 & 0.4750 & 0.0347 & 0.20\% & 0.3954 & 0.5981 & 0.0386 & 0.60\% & 0.2348 & 0.0641 & 0.0640 & 0.27\% \\
         CorrelNet & 0.6886 & 0.7573 & 0.0145 & 16.11\% & 0.5470 & 0.7873 & 0.0184 & 12.17\% & 0.2883 & 0.3948 & 0.0131 & 13.17\% \\
         \hline
         PF-AFN & 0.6403 & 0.7391 & 0.0186 & 7.86\% & 0.5325 & 0.7579 & 0.0235 & 6.27\% & 0.2735 & 0.2314 & 0.0201 & 1.43\% \\
         CP-VTON+ & 0.6596 & 0.7485 & 0.0159 & 8.01\% & 0.5321 & 0.7395 & 0.0349 & 4.91\% & 0.2758 & 0.1232 & 0.1989 & 0.67\% \\
         \hline
         Ours(w/o sm) & 0.6971 & 0.7753 & 0.0125 & 20.44\% & 0.5656 & 0.8181 & 0.0158 & 27.17\% & 0.3146 & 0.4744 & 0.0094 & 38.07\% \\ 
         Ours & \textbf{0.7038} & \textbf{0.7814} & \textbf{0.0122} & \textbf{31.98\%} & \textbf{0.5718} & \textbf{0.8239} & \textbf{0.0147} & \textbf{33.62\%} & \textbf{0.3229} & \textbf{0.4832} & \textbf{0.0093} & \textbf{43.20\%} \\
         \hline
    \end{tabular}}
    \caption{The quantitative results on STRAT dataset. The best results are emphasized in bold. “User” means the frequency that each method is chosen as the best method in user study.}
    \label{tab:eval}
\end{table*}

Although the main branch alone can predict the target locations of foreground keypoints, it is challenging to make precise predictions in the absence of foreground information. For example, in the hat try-on task, although the four target locations can be roughly positioned around the top of head, only with the specific shape and the type of hat (\emph{e.g.}, baseball cap, cowboy hat) can they be more precisely decided. Therefore, it is necessary to inject foreground information into the main branch.

We first concatenate the foreground image $I_f$ and its mask $M_f$, and feed them into the foreground encoder $E_f$ to extract the foreground feature map $F_f$. Then, we fuse the foreground feature map $F_f$ and background feature map  $F_b$ in the bottleneck to ensure that the heatmap decoder $D_{hm}$ could receive both foreground and background information. Inspired by previous works on attention~\cite{woo2018cbam} and feature fusion~\cite{dai2021attentional}, we propose a dual-attention fusion (DAF) module to fuse two feature maps attentively. The DAF module first applies dual attention (channel attention and spatial attention in \cite{woo2018cbam}) to the summation of two feature maps to predict the fusion weight map $M_{fuse}$ with the same size as $\{F_f, F_b\}$. Then, $F_f$ and $F_b$ are fused according to the fusion weight map $M_{fuse}$:
\begin{equation}
    F_{fuse} = M_{fuse}\otimes F_b + (1 - M_{fuse}) \otimes F_f,
\end{equation}
where $F_{fuse}$ is the fused feature. The values in $M_{fuse}$ are between 0 and 1 after Sigmoid, enabling our network to make a soft selection between two feature maps. The DAF module is illustrated in Figure \ref{fig:model}, in which the dashed line means reversing the fusion weight map.

\subsection{Utilizing Auxiliary Background Information}\label{sec:sm}
Due to our design of background-oriented network, it can be easily extended to utilize additional background information. In this work, we take semantic information as an example to investigate how to leverage auxiliary background information in our network. It is reasonable to presume that background semantic information could benefit target location hallucination, because the relative position and distance to different semantic components (\emph{e.g.}, eyes, nose, ears) could help determine the target locations of foreground keypoints. Similar to previous multitask learning works~\cite{tsai2017deep, wang2020multi, bhattacharjee2022mult, carvalho2019multitask}, we introduce a semantic decoder $D_{sm}$  to predict background semantic mask. The semantic decoder $D_{sm}$ and the heatmap decoder $D_{hm}$ share the same background encoder $E_b$. $D_{sm}$ and $E_b$ also have skip connections. 
To leverage the semantic knowledge learned by $D_{sm}$, the feature map of each layer in the decoder $D_{sm}$ is propagated to the decoder $D_{hm}$ by adding to the corresponding feature map, except the last layer which accounts for heatmap prediction. 

\paragraph{Loss Function}
We denote the predicted semantic mask as $\hat{S} \in \mathbb{R}^{C \times N}$, in which $C$ is the number of semantic classes (\emph{e.g.}, eyes, nose, ears) and $N$ is the size of semantic mask (the same as the size of heatmap). Besides, we use $c_p$ to denote the ground-truth label at location $p$. Then, the pixel-wise cross-entropy loss can be represented by
\begin{equation}
    \mathcal{L}_{sm} = -\frac{1}{N}\sum_{p=1}^N\log\left(\frac{\exp{\hat{S}_{c_p, p}}}{\sum_{c=1}^C \exp{\hat{S}_{c, p}}}\right).
\end{equation}

The overall objective function can be written as
\begin{equation}\label{eq:loss}
    \mathcal{L} = \mathcal{L}_{hm}+\lambda \mathcal{L}_{sm},
\end{equation}
where $\lambda$ is a hyper-parameter to balance two tasks.

\section{Experiments}
\subsection{Dataset and Evaluation Metrics}

We conduct experiments on STRAT dataset~\cite{zhang2022spatial}, which consists of three sub-datasets (STRAT-glasses, STRAT-hat, STRAT-tie). Each sub-dataset contains 2000 training tuples and 1000 test tuples. Each tuple has a foreground image with mask, a background image, and the ground-truth composite image.
The foreground/background images in the training set and test set have no overlap. For quantitative evaluation,  we adopt local structural similarity index (LSSIM), vertex displacement error (Disp), and intersection over union (IoU) following~\cite{zhang2022spatial}. We also consider human perception and include user study for more comprehensive comparison. Specifically, we collect the composite images generated by different methods for the whole test set. 20 human raters are asked to select the most reasonable and realistic result for each test tuple. Then, we report the frequency that each method is selected as the best one. 

\subsection{Implementation Details}
We use PyTorch~\cite{paszke2019pytorch} v1.11.0 to implement our model, and train it on the NVIDIA RTX 3090 GPU. In the STRAT-glasses and STRAT-hat sub-datasets, we resize both foreground and background images to $224\times 224$. For the STRAT-tie sub-dataset, we resize the images to $448 \times 448$ because the foreground object only occupies a small proportion of the background, following~\cite{zhang2022spatial}. We set the Gaussian radius of ground-truth heatmap as $g=20$, and set $\gamma=10$ in Eqn.~\ref{eq:wm} and $\lambda=0.1$ in Eqn.~\ref{eq:loss} via cross-validation.

We use adam optimizer with the learning rate initialized as 0.0002 and weight decay set to $1\times 10^{-5}$. The batch size is 32 and our model is trained for 40 epochs with the learning rate decreasing linearly from epoch 10 to 0.00005.

For the semantic classes of the background image, we divide three sub-datasets into two groups. The background images in the first group (STRAT-glasses and STRAT-hat) are human faces and those in the second group (STRAT-tie) are human bodies. 
Human face has 12 semantic classes: background, hair, hat, eyebrows, glasses, eyes, nose, mouth, skin, neck, ears, and clothing. Human body has 8 semantic classes: background, head, upper, lower, arms, legs, shoes, and bags.

\subsection{Comparison with Baselines}

\begin{figure*}[!ht]
\centering
\includegraphics[width=1.0\linewidth]{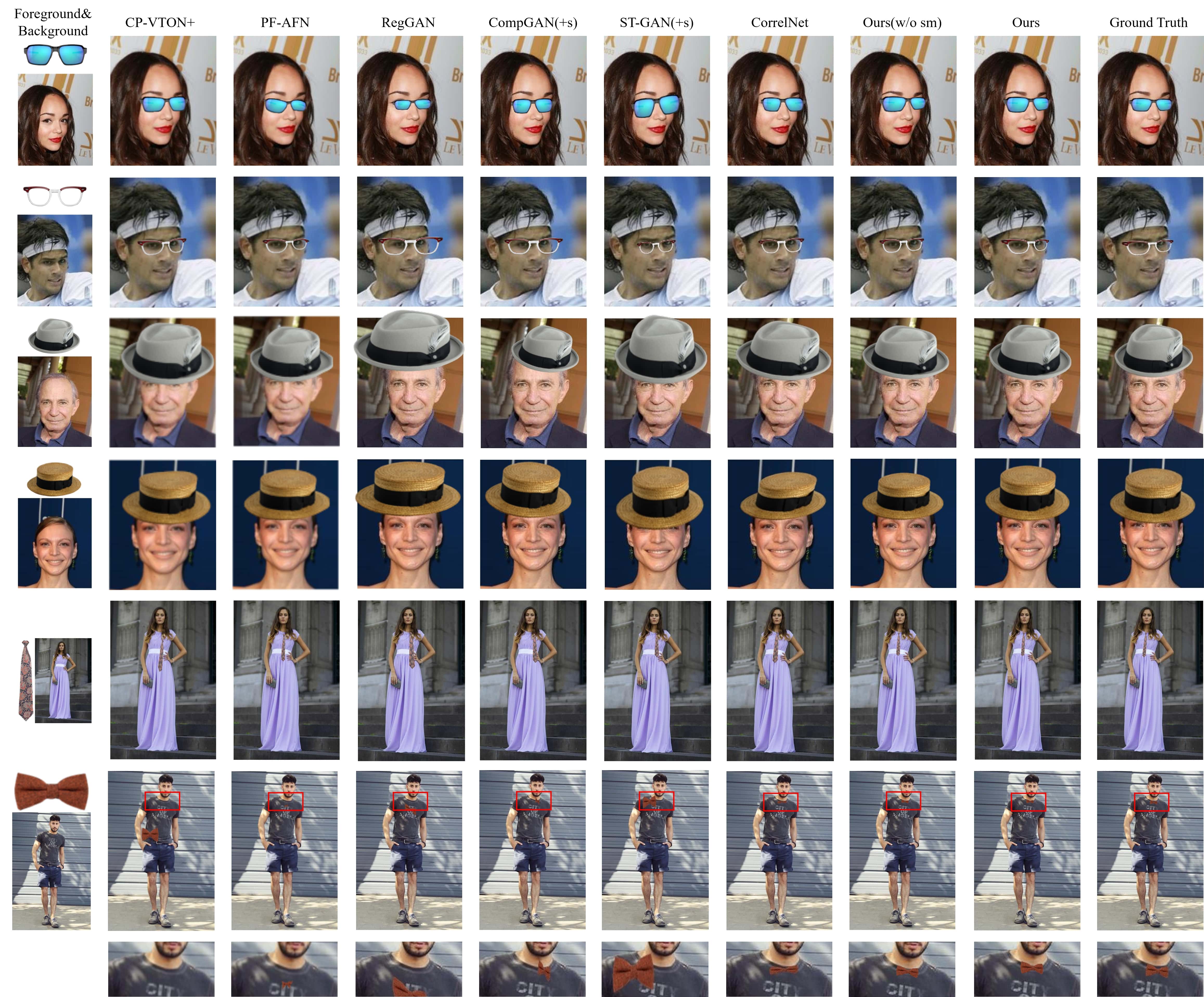}
\caption{Qualitative results of various methods on the STRAT dataset. In the last row, we zoom in to better observe the bow tie.}
    \label{fig:composite}   
\end{figure*}

We compare our method with previous virtual accessory try-on methods: CorrelNet~\cite{zhang2022spatial}, ST-GAN~\cite{lin2018st}, RegGAN~\cite{kikuchi2019regularized}, CompGAN~\cite{azadi2020compositional}, SF-GAN~\cite{zhan2019spatial}, AGCP~\cite{li2021image}, and representative virtual clothing try-on methods: CP-VTON+~\cite{minar2020cp}, PF-AFN~\cite{ge2021parser}.In particular, (+s)(\textit{e.g.}, ST-GAN(+s) and CompGAN(+s)) means adding additional supervision (MSE loss) between generated composite images and ground-truth composite images compared to their original methods, which allows them to utilize the ground-truth annotations more effectively (see~\cite{zhang2022spatial} for more details). Note that two virtual clothing try-on methods require semantic parsing, and we provide them with ground-truth semantic annotations. As for the other input of these two methods, \emph{i.e.} the pose heatmaps, we did not use the corresponding information into the input. And we only use the first stage of these two methods to obtain the warped foreground image, and then use the foreground mask as guidance to synthesize the two images, which is also consistent with the process of our method. In addition, since the optical flow-based warping method PF-AFN cannot accurately obtain the final locations of these specified vertices, we calculate the perspective transformation matrix by sampling some matching point pairs, through which we can obtain the final locations of there vertices to approximately estimate the Disp metric.

The quantitative results are summarized in Table \ref{tab:eval}. Among the virtual accessory try-on methods, the recent method CorrelNet~\cite{zhang2022spatial} achieves the best results, surpassing other methods by a large margin, which demonstrates the advantage of calculating warping parameters based on selected keypoint pairs. The virtual clothing try-on methods~\cite{minar2020cp, ge2021parser} are very competitive and outperform most early virtual accessory try-on methods.
We observe that~\cite{minar2020cp, ge2021parser} perform admirably on the STRAT-glasses and STRAT-hat datasets, but less ideally on the STRAT-tie dataset. The reason could be that with the foreground tie only occupying a small region on the background, the approach of seeking the global correspondence between foreground and background becomes less effective. This is more obvious in CP-VTON+ because it only establishes such correspondence on high-level features. Its poor performance on STRAT-tie dataset is also attributable to the difficulty in distinguishing between neck ties and bow ties. We refer to our method without semantic information as Ours(w/o sm). It can be seen that Ours(w/o sm) performs better than all previous methods on all metrics, which demonstrates the superiority of our method. After using semantic information, our approach achieves further improvement (Ours \emph{v.s.} Ours(w/o sm)).

\begin{table}[t]
    \centering
    \scalebox{0.92}{
    \begin{tabular}{c|ccc|ccc}
        \hline
          & SM & FG & $\mathcal{L}_{hm}$ & LSSIM$\uparrow$ & IoU$\uparrow$ & Disp$\downarrow$  \\
         \hline\hline
         1 &  &  & wAW & 0.4962 & 0.7323 & 0.0262 \\
         2 & + &  & wAW & 0.4979 & 0.7358 & 0.0258 \\
         3 &  & DAF & wAW & 0.5656 & 0.8181 & 0.0158 \\
         4 & + & add & wAW & 0.5678 & 0.8224 & 0.0149 \\
         5 & + & DAF(s) & wAW & 0.5701 & 0.8227 & 0.0149 \\
         6 & + & DAF & MSE & 0.5527 & 0.8165 & 0.0158 \\
         7 & + & DAF & AW & 0.5653 & 0.8211 & 0.0149 \\
         8 & + & DAF & wMSE & 0.5596 & 0.8217 & 0.0148 \\
         9 & + & DAF & wAW & \textbf{0.5718} & \textbf{0.8239} & \textbf{0.0147} \\
         \hline
    \end{tabular}}
    \caption{Ablation Studies of loss functions and network components of our model. AW (\emph{resp.}, wAW) means AWing (\emph{resp.}, weighted AWing) loss and MSE (\emph{resp.}, wMSE) means MSE (\emph{resp.}, weighted MSE) loss. FG (\emph{resp.}, SM) means foreground (\emph{resp.}, semantic).}
    \label{tab:ablation}
\end{table}

\subsection{Ablation Studies}
By taking STRAT-hat as an example, we conduct ablation studies to validate the effectiveness of each component in our network, and the results are shown in Table \ref{tab:ablation}. First, we only use the main branch (background encoder $E_b$ and heatmap decoder $D_{hm}$). As shown in row 1, the overall results are very poor. On the basis of row 1, we introduce semantic mask and foreground information respectively, and the results are shown in row 2 and row 3 respectively. As can be seen, the additional semantic information brings limited improvement in this case. The addition of foreground information results in substantial improvement, because our model is unable to predict heatmaps accurately without knowing the foreground information (\emph{e.g.}, shape and sub-type of hat).
On the basis of row 3, we add the semantic mask to arrive at our full-fledged method in row 9. 

Then, we investigate the strategy of injecting foreground information. One simple strategy is adding the feature maps {$F_b, F_f$} as fused feature map $F_{fuse}$. The obtained results in row 4 are lower than in row 9, which demonstrates that the DAF module can fuse foreground and background feature maps more effectively. Besides, we also experiment with a simplified version of DAF, which is referred to as DAF(s) in row 5. Specifically, we remove the dual attention~\cite{woo2018cbam} and directly output the fusion weight map. Compared with row 9, the features refined by dual attention are more conducive to feature fusion.

Next, we study the impact of various loss functions. We only use MSE loss or AWing loss~\cite{wang2019adaptive} to supervise the heatmap, as shown in row 6 and row 7 respectively. We observe that AWing loss is more superior in supervising heatmap prediction. In row 8 (\emph{resp.}, 9), we add weighted loss based on row 6 (\emph{resp.}, 7). Whether using MSE loss or AWing loss, assigning higher weights to the positive locations is helpful to address the imbalance issue. 

\subsection{Qualitative Analysis}
\paragraph{Composite Results}
The composite images produced by various methods on three sub-datasets of STRAT are exhibited in Figure \ref{fig:composite}. As can be seen, the composite images generated by many previous methods are irrational, whereas our results are more realistic and plausible, as well as closer to the ground-truth. Although the two virtual clothing try-on methods~\cite{minar2020cp, ge2021parser} perform well on the STRAT-glasses and STRAT-hat datasets, some illogical deformations may occur due to incompatibility of their warping methods, as illustrated in the first row second column. Besides, these two methods challenge handling such tiny accessories while coping with the STRAT-tie dataset. Particularly for the bow tie, they are tough to predict the size and location of the bow tie, which leads to poor results.

As for the previous methods of virtual accessory try-on, CorrelNet~\cite{zhang2022spatial} is undoubtedly the best performing one, the composite results of the other methods are not ideal. Compared with CorrelNet, our method still performs better in many aspects. Specifically, in the first two rows, compared to CorrelNet, the warped foreground of our method is more in line with the orientation of the face. Additionally, in the second row, our model can more accurately predict the tilt of the glasses with the help of semantic information. The middle two rows show that our method can estimate the size and orientation of hat better than CorrelNet, and can predict the size of hat better when semantic information is included. Similarly, in the tie try-on examples, our model performs better for predicting the size and shape of the tie. More examples of composite results and the discussion on limitation of our method are presented in the Supplementary.

\paragraph{Heatmap Visualization}
\begin{figure}[t]
    \centering
    \includegraphics[width=1.0\linewidth]{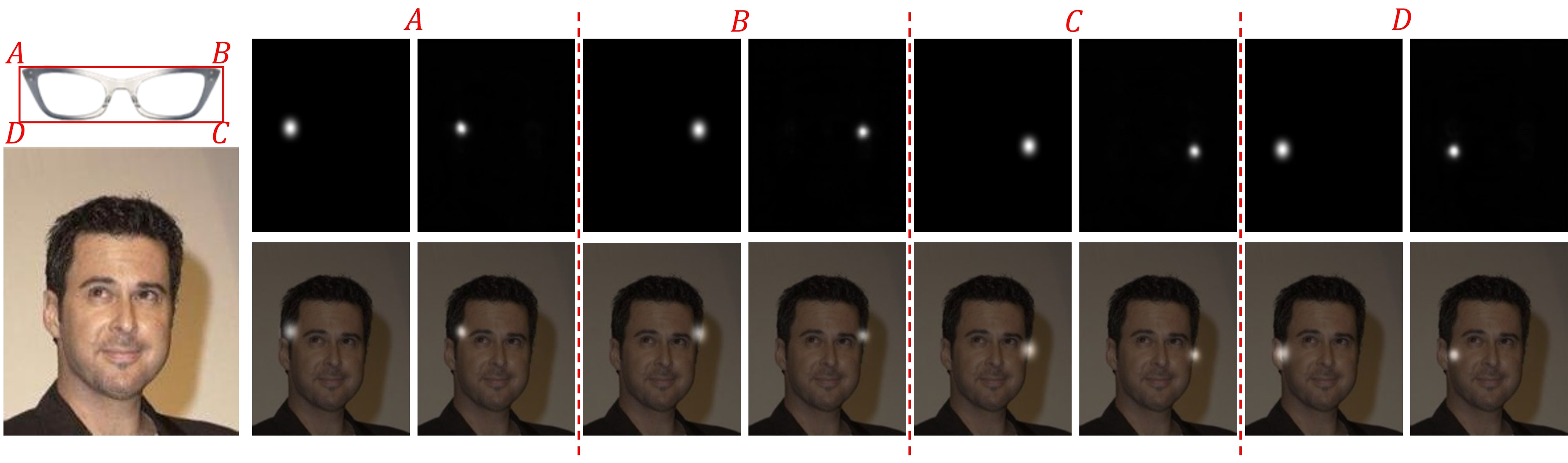}
    \caption{Visualization of heatmaps. On the left, we show the foreground with four keypoints and the background. In four groups corresponding to four foreground keypoints, we show the ground-truth heatmap in the left column and the predicted heatmap in the right column. We also superimpose the heatmap on the background for better visualization at the bottom.}
    \label{fig:heatmap}
\end{figure}
We visualize the ground-truth heatmaps and our predicted heatmaps for four foreground keypoints (A, B, C, D) in Figure \ref{fig:heatmap}. Our predicted heatmaps have Gaussian-like distribution and are basically consistent with the ground-truth heatmaps. The bright spots on our generated heatmaps are clearly visible and the overall heatmaps are not blurred, which can help us locate the predicted keypoints more accurately. 
Similarly, more visualization examples of heatmaps are included in Supplementary.
\paragraph{Semantic Mask}
\begin{figure}[t]
    \centering
    \includegraphics[width=1.0\linewidth]{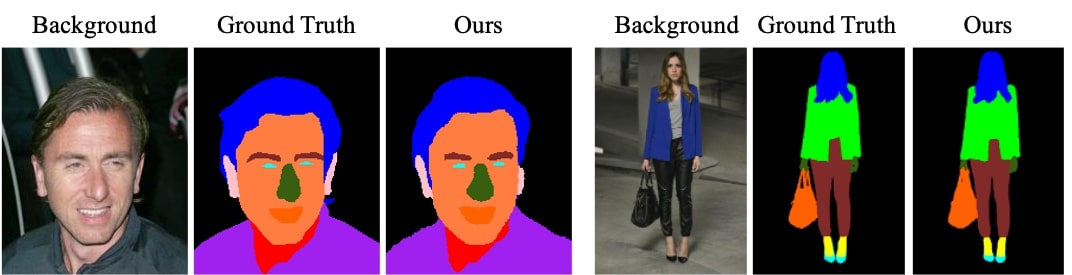}
    \caption{Visualization of semantic masks. In each example, we show the background image, the ground-truth semantic mask, and the predicted semantic mask (Best viewed in color).}
    \label{fig:mask}
\end{figure}
To verify that our model has learned useful semantic knowledge to help predict heatmaps, we visualize the semantic masks predicted by our model in Figure \ref{fig:mask}. Recall that human face images are used in the STRAT-glasses and STRAT-hat datasets, while human body images are used in the STRAT-tie dataset. Therefore, we show one example for each type of background images. The results demonstrate that, despite some inaccurate details, our model can learn the rough semantic information, which is qualified to help predict better heatmaps. 
For our tasks, such as hat try-on task, as long as our model can predict some semantic components (\emph{e.g.}, hair, ears, and facial skin) of people in the background, we can obtain a rough outline of the head, which can be helpful for keypoint heatmap generation.

\subsection{Hyper-parameter Analyses and Significance Test}
To better test the performance of our model, we perform hyper-parameter analyses and significance test. In the hyper-parameter analyses, we vary the values of three hyper-parameters (\emph{i.e.}, $\lambda$ in Eqn. \ref{eq:loss}, $\gamma$ in Eqn. \ref{eq:wm}, and the Gaussian radius $g$ of ground-truth heatmap), and plot the performance variance of our method to observe their impact. The significance test is conducted between our method and the strongest baseline CorrelNet~\cite{zhang2022spatial}. The detailed results are left to Supplementary due to the space limitation. 

\section{Conclusion}
In this work, we have proposed a novel way to address the virtual accessory try-on task, by converting the spatial transformation task to a keypoint hallucination task, that is, forecasting the target locations of foreground keypoints on the background.
We have designed a novel background-oriented network and explored using auxiliary background semantic information for performance improvement. The experimental results on the STRAT dataset have demonstrated the superiority of our method.

{\small
\bibliographystyle{ieee_fullname}
\bibliography{egbib}
}

\end{document}


\title{Supplementary for Virtual Accessory Try-On via Keypoint Hallucination}

\author{Junhong Gou\textsuperscript{1}, Bo Zhang\textsuperscript{1}, Li Niu\textsuperscript{1}, Jianfu Zhang\textsuperscript{1}, Jianlou Si\textsuperscript{2}, Chen Qian\textsuperscript{2}, Liqing Zhang\textsuperscript{1}\\
\textsuperscript{1}Shanghai Jiao Tong University, \textsuperscript{2}Sensetime Research\\
{\tt\small \{goujunhong, bo-zhang, ustcnewly, c.sis\}@sjtu.edu.cn, zhang-lq@cs.sjtu.edu.cn}\\
{\tt\small \{sijianlou, qianchen\}@sensetime.com}
}

\maketitle
\ificcvfinal\thispagestyle{empty}\fi

In this document, we provide additional materials to supplement our main text. In Section \ref{sec:hyper}, we study the effect of three hyper-parameters adopted in our work.  In Section \ref{sec:result}, we show more qualitative results on the STRAT dataset. In Section \ref{sec:sig}, we perform significance test. Additionally, we show some failure cases generated by our method and discuss the limitation of our method in Section \ref{sec:limit}.
\section{Hyper-parameter Analyses}\label{sec:hyper}

\begin{figure*}[t]
	\centering
	\includegraphics[width=1.0\linewidth]{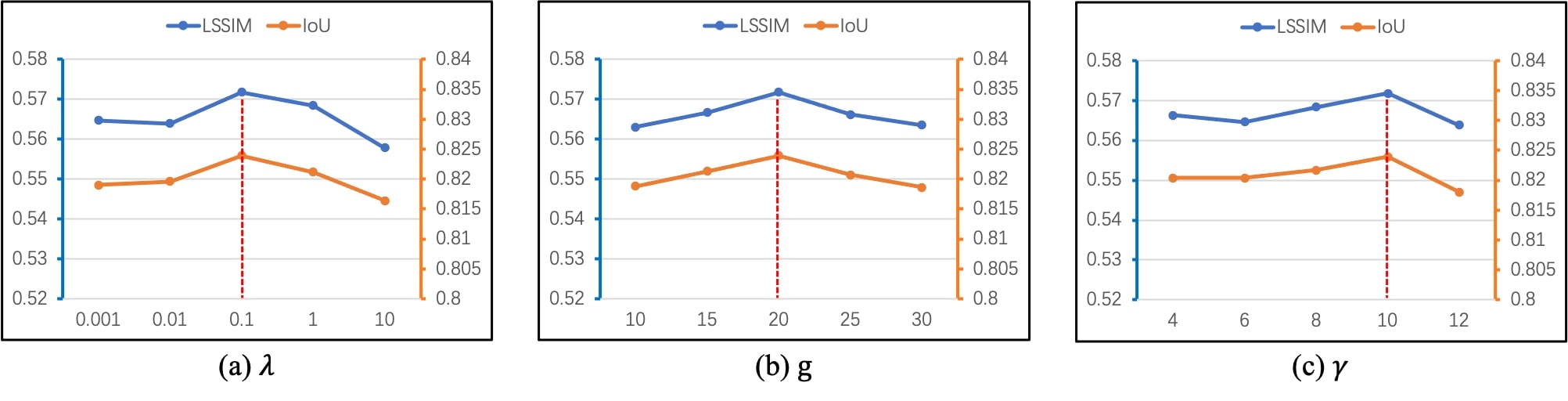}
	\caption{Performance variation of our method with different hyper-parameters $\lambda$ in Eqn. 8, $g$ (Gaussian radius of ground-truth heatmap) and $\gamma$ in Eqn. 3 in STRAT-hat sub-dataset. The red vertical dashed line represents the default value used in our other experiments.}
	\label{fig:hyper}
\end{figure*}

Recall that in the main text, we have a hyper-parameter $\lambda$ in Eqn. 8, which is set as $0.1$ via cross-validation. In addition, $\gamma$ in Eqn. 3 and the Gaussian radius $g$ are set to $20$ and $10$ respectively. By taking STRAT-hat as an example, we further plot the performance variance of our method when varying these hyper-parameters. 

For $\lambda$, we vary it in the range of $[0.001,10]$ and plot the results in Figure \ref{fig:hyper} (a). When $\lambda=0.1$, both LSSIM and IoU reach the maximum value. The proper range of $\lambda$ is $[0.1, 1]$, and the value of $\lambda$ that are too large or too small can significantly affect performance. Then, we vary $g$ in the range of $[10, 30]$ and show the results in Figure \ref{fig:hyper} (b). It is obvious that our model achieves the best performance when $g=20$. When $g$ is in $[15, 25]$, the IoU of the model is between $[0.8207, 0.8239]$ and the LSSIM is between $[0.5662, 0.5718]$, which indicates that our model is robust when the value of $g$ is within a reasonable range. With $g=20$, we experiment with $\gamma$ in the range of $[4, 12]$. As shown in Figure \ref{fig:hyper} (c), $\gamma=10$ reaches the maximum value and there is a significant drop in performance when $\gamma > 10$. Therefore, if we give too much weight to the positive locations defined in Section 3.2 of the main text, the model will fail to converge, while the imbalance issue is  not very well solved with less weight.

\section{More Qualitative Results}\label{sec:result}
\subsection{Composite Results}
\begin{figure*}[h]
    \centering
    \includegraphics[width=1.0\linewidth]{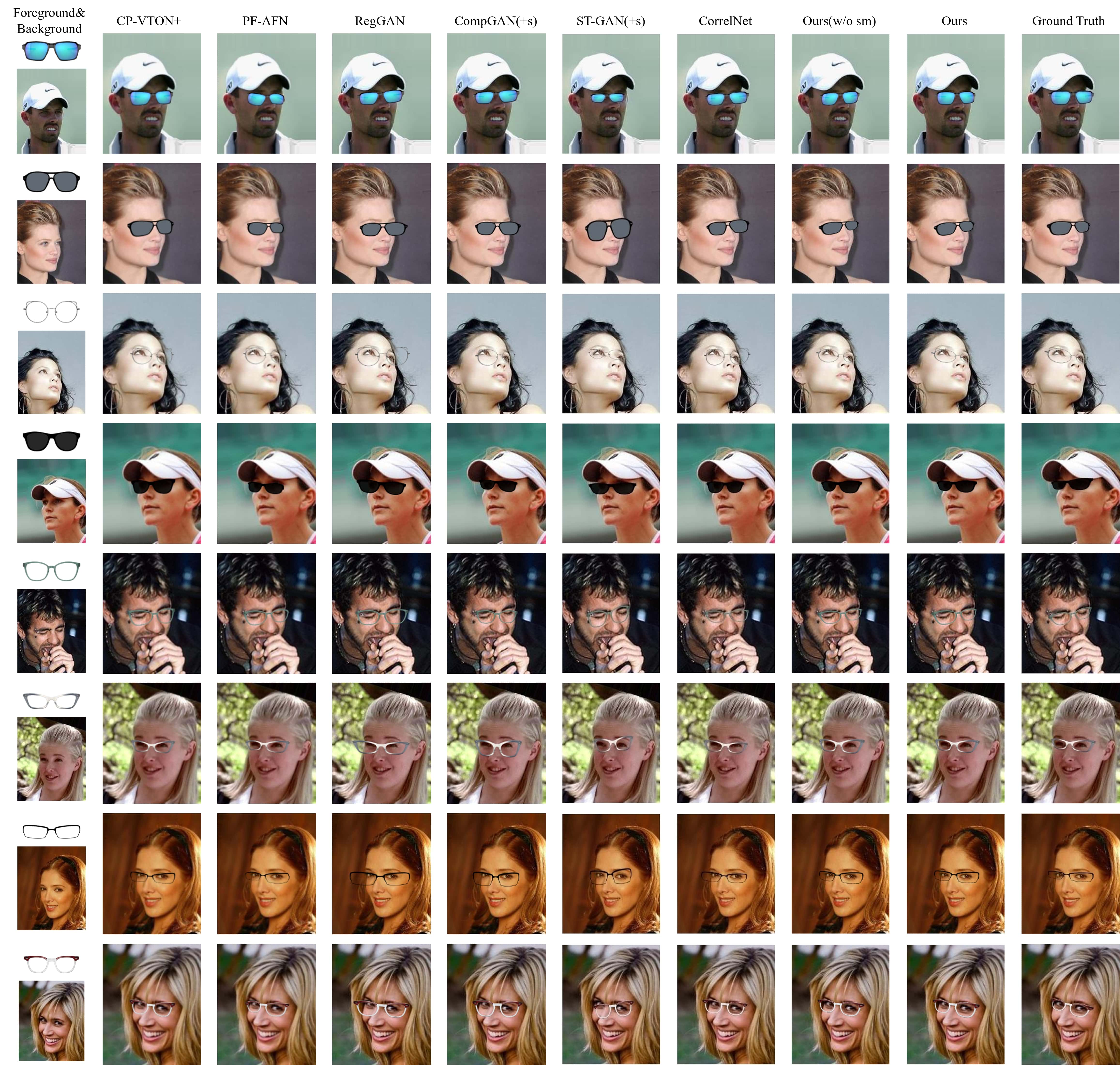}
    \caption{Qualitative comparison of different methods on STRAT-glasses sub-dataset.}
    \label{fig:glasses}
\end{figure*}

\begin{figure*}[h]
    \centering
    \includegraphics[width=1.0\linewidth]{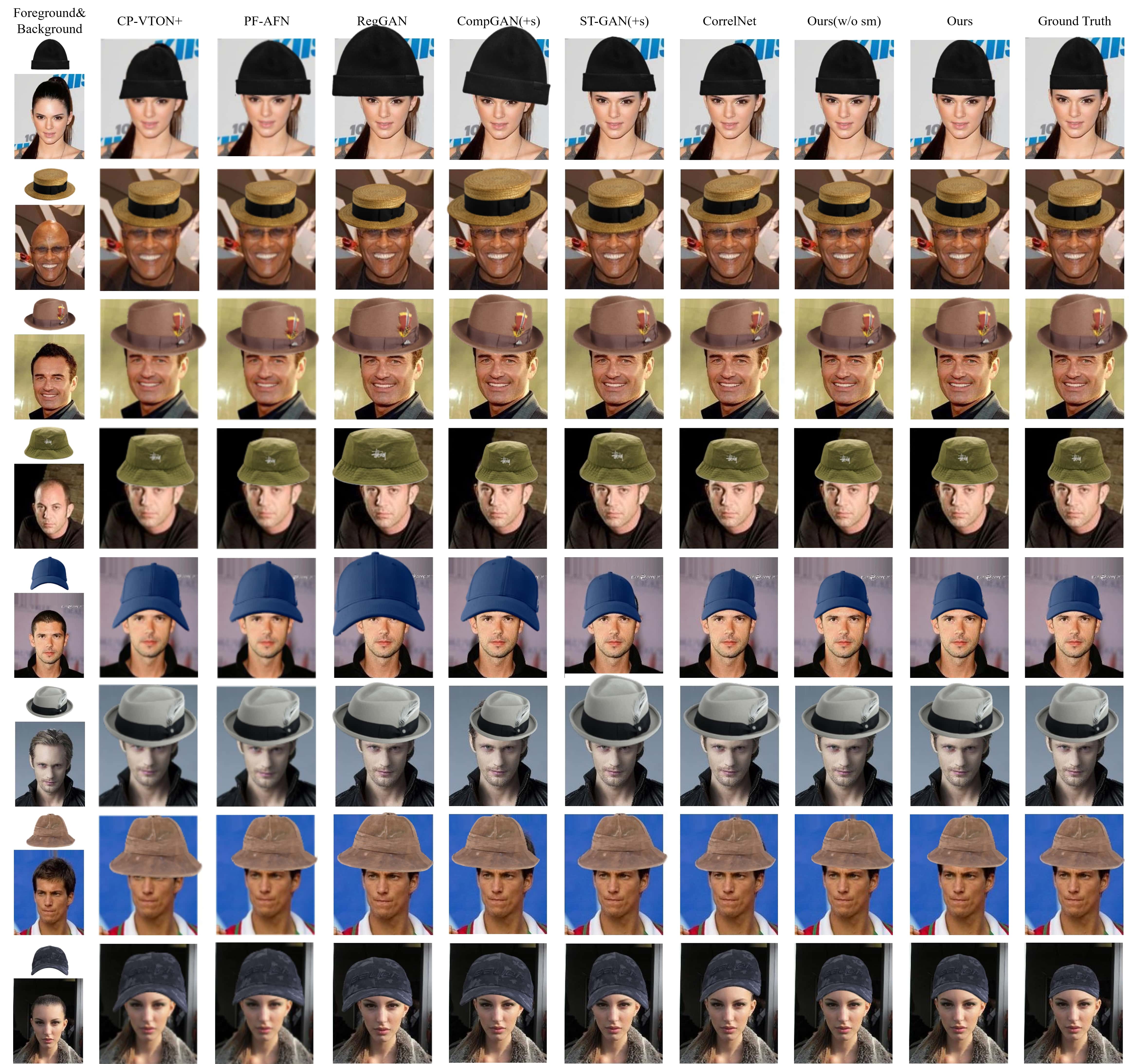}
    \caption{Qualitative comparison of different methods on STRAT-hat sub-dataset.}
    \label{fig:hat}
\end{figure*}

\begin{figure*}[h]
    \centering
    \includegraphics[width=1.0\linewidth]{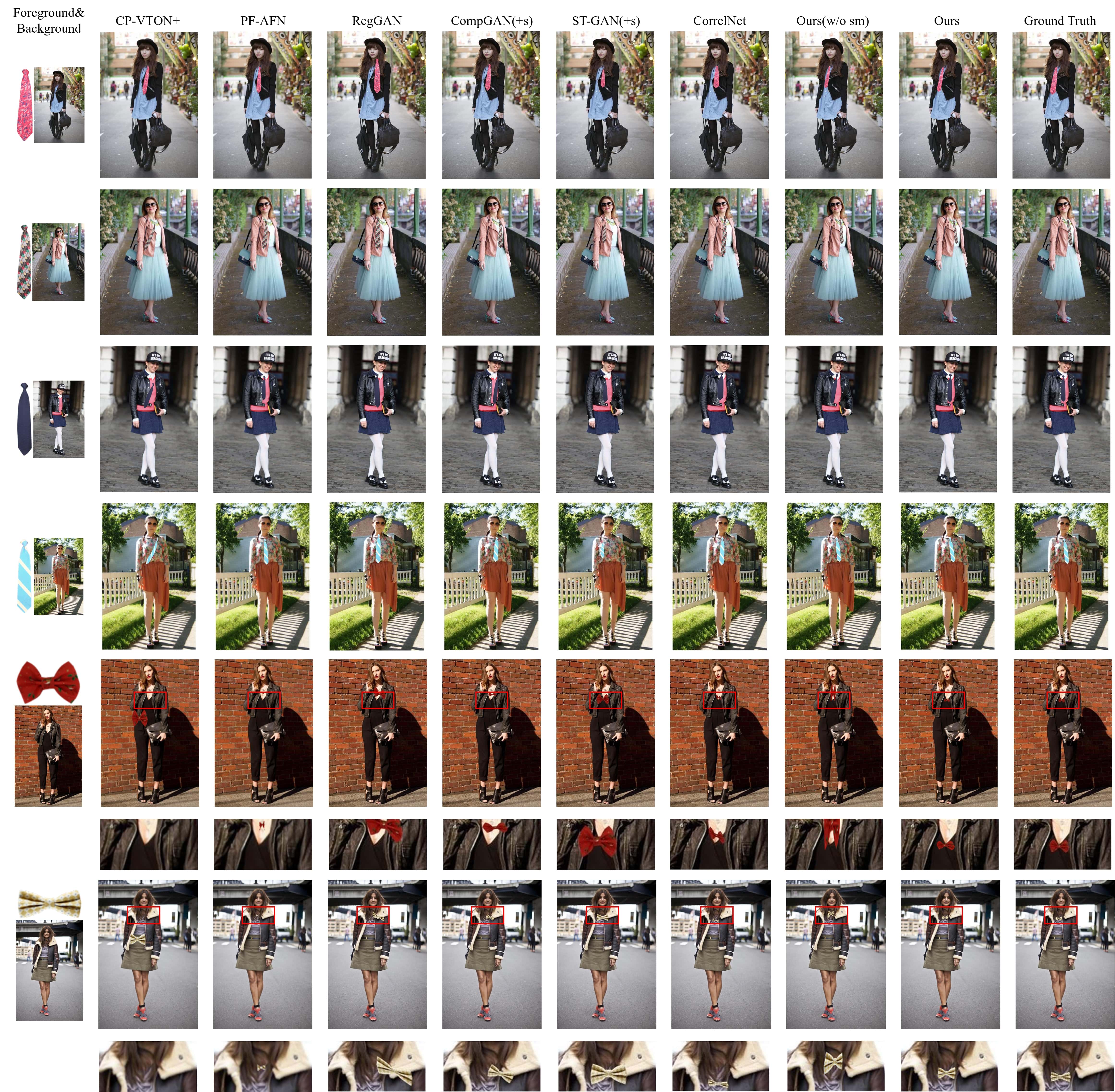}
    \caption{Qualitative comparison of different methods on STRAT-tie subdataset. In the last two rows, we zoom in for better observation.}
    \label{fig:tie}
\end{figure*}

In this section, we show more composite images produced by various methods on three subdatasets of STRAT~\cite{zhang2022spatial}, \emph{i.e.}, STRAT-glasses, STRAT-hat, STRAT-tie, in Figure \ref{fig:glasses}, Figure \ref{fig:hat}, Figure \ref{fig:tie}, respectively. The previous methods include CP-VTON+~\cite{minar2020cp}, PF-AFN~\cite{ge2021parser}, RegGAN~\cite{kikuchi2019regularized}, CompGAN~\cite{azadi2020compositional}, ST-GAN~\cite{lin2018st}, and CorrelNet~\cite{zhang2022spatial}. 
It can be seen that our method can produce more realistic and plausible composite results for a variety of foreground and background. .pngIn contrast, the placement of the foreground in the composite images generated by many previous methods is not very reasonable.
For the virtual clothing try-on methods~\cite{minar2020cp, ge2021parser}, although they can produce competitive results on the STRAT-glasses and STRAT-hat datasets, they perform poorly on the STRAT-tie dataset. In row 1 and 2 of Figure \ref{fig:glasses}, we can see that these two methods produce some unreasonable deformations, which results in unrealistic synthetic results. Meanwhile, in the case of some faces with large rotations, these two methods cannot be very good. Besides, these two methods challenge to deal with some situations where the face has a large rotation. For hat try-on task, as shown in Figure \ref{fig:hat}, both methods result in some unnatural distortion like row 5 and 8, and have difficulty in warping the hat to the proper size. 
In Figure \ref{fig:tie}, the neck ties warped by CP-VTON+ are unnaturally twisted, and the bow ties are also not in the correct zone. PF-AFN is unable to place neck ties to proper position when persons are slightly sideways, also unable to warp bow ties to proper size. 

For previous virtual accessory try-on methods, most of them fail to produce realistic results. CorrelNet is the strongest baseline method, and we will also focus on comparing with it.
Specifically, in Figure \ref{fig:glasses}, since many faces have a certain tilt, the glasses warped by CorrelNet in the first four rows do not conform to the orientation of the face well. However, our method can generate glasses that are basically aligned with the face orientation even without semantic information (denoted as Ours(w/o sm)). With the semantic information, our method can better predict the size of the glasses, and also improve the results of some samples that cannot be handled well by Ours(w/o sm), such as the first row. In Figure \ref{fig:hat}, in many examples, the hat orientation of the composite image of CorrelNet is inconsistent with the hat orientation of the ground-truth composite image, such as rows 1, 2, and 5. In contrast, our method works better. In some examples, the predicted hat size is more suitable by incorporating semantic information, such as rows 2, 3, and 6. For the tie try-on task in Figure \ref{fig:tie}, our method outperforms other methods. Compared to CorrelNet, the results predicted by our method for neckties are closer to ground-truth in size and orientation. Semantic information is of great help in the prediction of necktie size. For the bowtie, the position of the bowtie is more accurately predicted with the help of semantic information.

\begin{figure}[ht]
    \centering
    \includegraphics[width=1.0\linewidth]{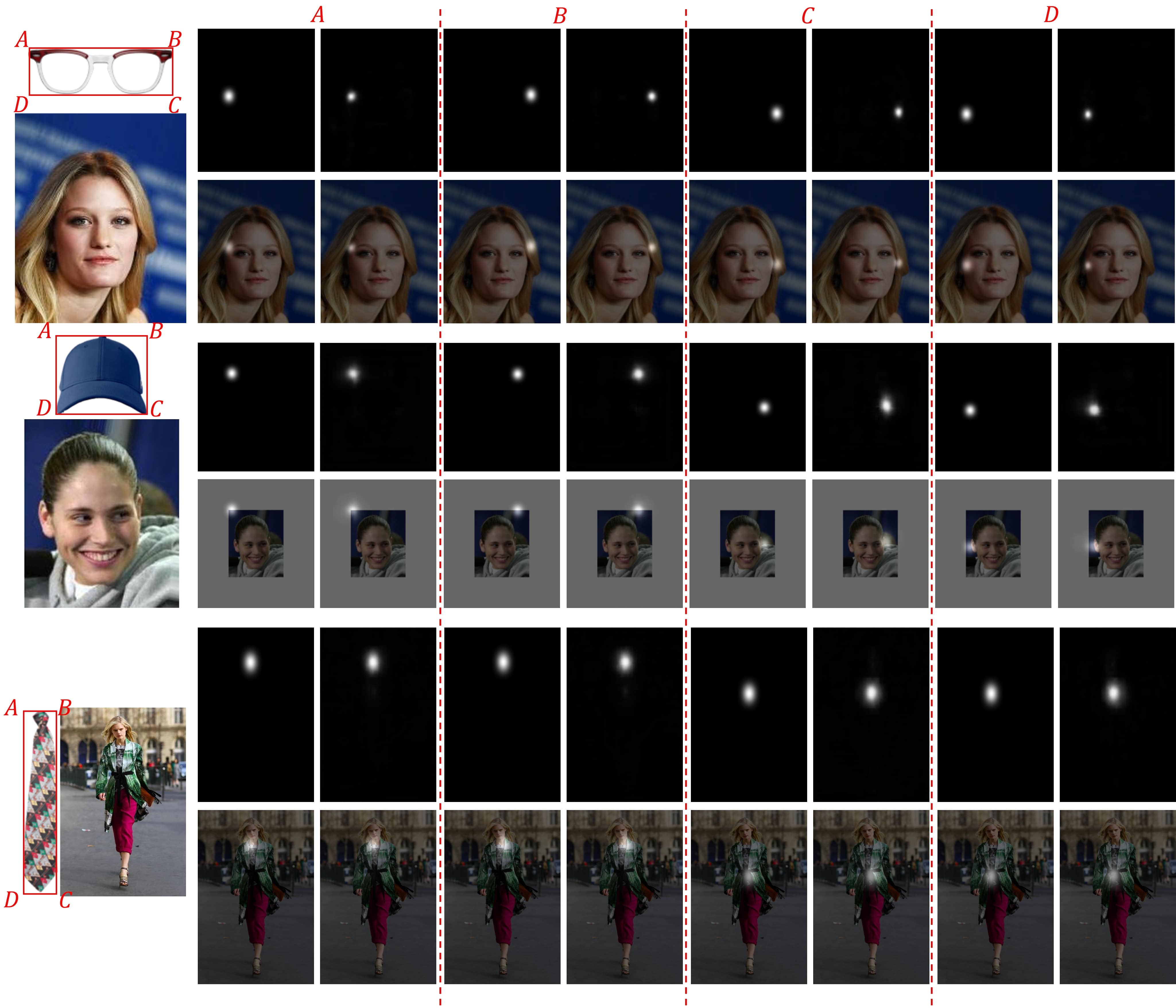}
    \caption{Visualization of heatmaps. The left shows the background and foreground with four keypoints. In the four groups corresponding to keypoints, the left column of each group shows the ground-truth heatmap, and the right column is the predicted heatmap. At the bottom of each example, we superimpose the heatmap on the background.}
    \label{fig:heatmaps}
\end{figure}

\begin{figure}[ht]
	\centering
	\includegraphics[width=1.0\linewidth]{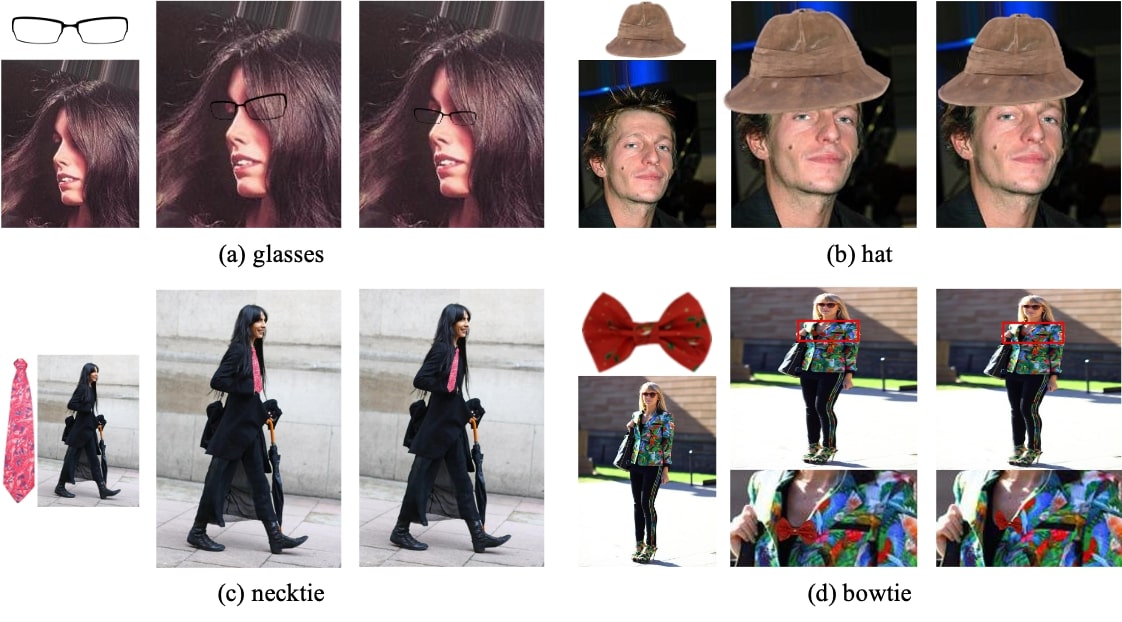}
	\caption{Visualization of our failure cases on STRAT dataset. In each example, from left to right: foreground and background images, ground-truth composite image, and our generated composite image. For (d) bowtie, we zoom in for clearer observation.}
	\label{fig:fail}
\end{figure}


\subsection{Heatmap Visualization}
We select one example from each of the three sub-datasets to visualize the ground-truth and predicted heatmaps, as shown in Figure \ref{fig:heatmaps}. We divide these heatmaps into four groups corresponding to four foreground keypoints (A, B, C, D). It is evident that the highlighted areas of the predicted heatmaps and the ground-truth heatmaps are highly similar. For different foregrounds, the location of the bright spot matches the expected target location. For example, these four keypoints fall around the eyes for glasses, around the head for hats, and around the neckline and waist for ties. The injection of foreground information helps our model locate these keypoints in the background more accurately to obtain a more realistic composite image.

\section{Significance Test}\label{sec:sig}
We perform the significance test between our proposed method and the strongest baseline CorrelNet~\cite{zhang2022spatial}. By taking STRAT-hat dataset as an example, we run our method and CorrelNet 10 times with random seeds ranging from 1 to 10. The LSSIM, IoU, and Disp results of our method are 56.69 $\pm$ 0.45$\%$, 82.05 $\pm$ 0.34$\%$, and 1.51 $\pm$ 0.05$\%$, while the results of CorrelNet are 54.44 $\pm$ 0.52$\%$, 78.49 $\pm$ 0.38$\%$ and 1.82 $\pm$ 0.06$\%$. At the significance level of 0.05, we perform Welch's T-Test to verify that our method is better than CorrelNet. The p-value is much smaller than 0.001, which demonstrates that the superiority of our method is statistically significant.

\section{Discussions on Limitation}\label{sec:limit}
Despite producing excellent results, our method does not entirely cover all cases. As shown in Figure \ref{fig:fail}, we display some less satisfactory composite results. Our method does not produce a good composite result in Figure \ref{fig:fail} (a), because the background around the eyes is obscured and our method needs to predict the target locations around the eyes. In Figure \ref{fig:fail} (b), our method places the hat at a higher position and incorrectly estimates the size of hats, which may be due to the large tilt angle of human faces. In both \ref{fig:fail} (c) and \ref{fig:fail} (d), the foreground objects are not warped to the correct angle. One possible explanation is that the foreground person is not facing the front and our model misjudges the orientation of human bodies. For a tie, since the foreground accounts for a small proportion of the background, this may also be an important factor that leads to the inability to accurately predict the orientation and size of it.

{\small
\bibliographystyle{ieee_fullname}
\bibliography{egbib}
}